\documentclass[authorversion,nonacm]{acmart}
\usepackage{algorithm} 
\usepackage{algpseudocode}
\usepackage{tikz}
\usepackage{multirow}
\usepackage{gensymb}
\AtBeginDocument{%
  }

\begin{document}

\title{Speaking images. A novel framework for the automated self-description of artworks.}

\author{Valentine Bernasconi}
\email{valentine.bernasconi@unibo.it}
\orcid{0000-0002-9467-8896}
\authornotemark[1]
\affiliation{%
  \institution{University of Bologna}
  \city{Bologna}
  \country{Italy}
}

\author{Gustavo Marfia}
\email{gustavo.marfia@unibo.it}
\affiliation{%
  \institution{University of Bologna}
  \city{Bologna}
  \country{Italy}
}

\renewcommand{\shortauthors}{Bernasconi et al.}

\begin{abstract}
Recent breakthroughs in generative AI have opened the door to new research perspectives in the domain of art and cultural heritage, where a large number of artifacts have been digitized. There is a need for innovation to ease the access and highlight the content of digital collections. Such innovations develop into creative explorations of the digital image in relation to its malleability and contemporary interpretation, in confrontation to the original historical object. Based on the concept of the autonomous image, we propose a new framework towards the production of self-explaining cultural artifacts using open-source large-language, face detection, text-to-speech and audio-to-animation models. The goal is to start from a digitized artwork and to automatically assemble a short video of the latter where the main character animates to explain its content. The whole process questions cultural biases encapsulated in large-language models, the potential of digital images and deepfakes of artworks for educational purposes, along with concerns of the field of art history regarding such creative diversions.
\end{abstract}




\maketitle

\section{Introduction}
Images are meant to convey messages. They are an expression of the world, a word, a metaphor, or the mythical. Especially in the case of religious images, such as Christian paintings, they showcase invisible worlds whose meaning often relies on the quality of looking of the spectator \citep{drury_painting_1999}. As much as the work of the painter is based on training and tradition, the gaze of the viewer is shaped by its cultural surroundings, emotional state, and mental image and becomes an instance of narrative coherence \citep{bauer_beyond_2017}. Yet, the image belongs to the sensible world and is subject to historical time \citep{strine_image_2022}. Cultures, doctrines, references, and representations evolve, as well as the object itself, whose tangibility recently shifted towards a dematerialized representation in the digital space. The original function of these images evolved from liturgical to ornamental purposes, from medieval complex systems meant to shape animated oracular speech through magic, miracle or mechanics \citep{jorgensen_animation_2023}, to still images. Hence, the meaning of the depiction is often overlooked by the modern public. Unless the viewer benefits from a proper training in history, art history or theology, the reading of these images can be quite difficult as references are missing. As explained by George Didi-Huberman, before an image, however old or recent, the present never ceases to reshape it, as it becomes a product of the memory \citep{farago_before_2003}. We therefore attempt to address in the present work the role of the digital image and its new modes of communication with the contemporary and inexperienced eye, changes in interpretive frameworks, the timelessness nature of the image \citep{staff_painting_2022}, and its dematerialization.

In recent years, the development of large language models (LLMs) and advancements in computer vision have opened the door to new research tracks in relation to LLM-based conversational agents such as ChatGPT\footnote{\url{https://chatgpt.com/gpts}}, or text-to-image generators, with for instance DALL-E\footnote{\url{https://openai.com/index/dall-e/}}, Midjourney\footnote{\url{https://www.midjourney.com/}} or Stable Diffusion\footnote{\url{https://www.diffus.me/}}. On the one hand, researchers have attempted to understand the content of Internet-scaled training datasets used to produce such AI-generative models. Because they are trained with large amounts of image-text pairs, the models learn relations between visual content and their description in natural language that inevitably lead to the encapsulation of cultural biases in their latent space \citep{palmini2024patternscreativityuserinput, alkhamissi2024investigatingculturalalignmentlarge, li2024cultureparkboostingcrossculturalunderstanding, prabhakaran2022culturalincongruenciesartificialintelligence, 10.1145/3593013.3594095, 10378621}. On the other hand, these models fostered the development of new generative ones such as Image-to-Text \citep{wang2022gitgenerativeimagetotexttransformer, 10.1145/3581783.3611891, QIN2023120706, DANESHFAR2024109288}, which offer a wide range of use cases for image captioning in large datasets, Text-to-Speech \citep{casanova2024xttsmassivelymultilingualzeroshot, chen2024valle2neuralcodec, 10445948, wang2025sparkttsefficientllmbasedtexttospeech, 10852161} and, ultimately, Text-to-Video and Image-to-Video generators \citep{10887583, 9439899}, described as important breakthroughs for our society \citep{NEURIPS2024_c6483c8a}. For example, the model Sora by OpenAI\footnote{\url{https://openai.com/sora/}}, which is trained to generate videos of realistic or imaginative scenes from text prompts, seems to surpass previous attempts in the creation of Text-to-Video generators \citep{sun2024soraseesurveytexttovideo, sanchez-acedo_influence_2024}. In parallel, the field of face animation based on video or audio input also benefits from an important hype in relation to deepfakes \citep{lee2024tugofwardeepfakegenerationdetection, Rehaan02012024, mihailova_dally_2021, guo2025liveportraitefficientportraitanimation} and new concepts such as virtual interactions and digital doubles \citep{cilchis-survey-2023, 10.1145/3527850, Zhang_2023_CVPR, wei2024aniportraitaudiodrivensynthesisphotorealistic, xu2024hallohierarchicalaudiodrivenvisual}. Overall, these different technological advancements represent major progress in the field of computer vision and help the production of new artworks. Indeed, used in a great variety of domains, generators seem to be particularly attractive for media and creative industries, as well as education. They open debates on different notions such as creativity, originality, and authorship \citep{10.1145/3600211.3604681, Madhu2025-MADAAA-6, 10937995, wu-ai-divencreativity2025, sanchez-acedo_influence_2024} along with issues on reliability, accuracy, and plagiarism \citep{Farrokhnia03052024, baidoo2023education, dwivedi2023opinion, 10.3389/bjbs.2024.14048, 10.1145/3613904.3642731} and force the creation of so-called guardrail systems to avoid the production of harmful contents \citep{dong2024buildingguardrailslargelanguage, dong2024safeguardinglargelanguagemodels}. Furthermore, artistic objects, such as paintings, or sculptures, still portray complex and challenging contents for the application of different detection and generative techniques, such as face and pose recognition \citep{bengamra-surveydetectionart2024, bernasconi_computational_2023, ju_human-art_2023}.

Detaching from classic historical practices based on the manual study and collection of historical sources to attribute an artwork, the present work aims to shape a new vision of the digitized artifact. Similarly to emerging works on automated descriptions of artworks through LLMs \citep{khadangi2025cognartivelargelanguagemodels, hayashi2024artworkexplanationlargescalevision, Bin_2024}, the understanding of the image results from the encapsulation of a concentrated western culture of the LLM \citep{Bin_2024, palmini2024patternscreativityuserinput, 10.1145/3593013.3594095, 10378621}. The scope goes beyond primitive Warburgian approaches based on the compilation of multiple sources and documents to explain influences, towards the genesis of an artwork \citep{bonneau_lire_2022}. Starting from the concept of the autonomous image, the artwork literally starts speaking for and about itself. More specifically, the paper proposes a pipeline for the automated description of any artwork based on a text produced by an LLM, face detection and animation. The result is a video of the animated artwork, hereafter called the speaking image, where depicted characters come to life to explain the content of the artwork. Built on top of recent advanced generative models, and on the recent hype about deepfakes, the process aims to showcase the potential behind their concatenation and to establish a critical perspective on their use in the cultural domain. Although the method seems very modern, the animation of the image has long been practiced and discussed. As showcased in recent medieval studies, different concepts of animations, from the image as a mechanism, to an organism, and to a social agent, have been practiced since medieval time \citep{jorgensen_animation_2023}. Transformations of the speaking image insert into the continuity of a discourse about the presence of the image and its social function.

In this work, we first present the whole process towards the production of the speaking image. We then introduce the different models used to assemble the pipeline, as well as the algorithms to combine the animated face with the whole original picture and the audio\footnote{The code is made publicly available on the Github repository \url{https://github.com/VBernasconi/Speaking-Images}}. We explain their selection and application based on performance criteria on artworks and the quality of the desired output. Finally, we discuss different issues introduced by the use of LLMs for artwork description, possible enhancements of such models with respect to historical rigor, as well as benefits from the perspective of automated handling of large cultural datasets.

\section{Materials and Methods}
\subsection{The overall process}
\begin{figure}[H]
\includegraphics[width=15 cm]{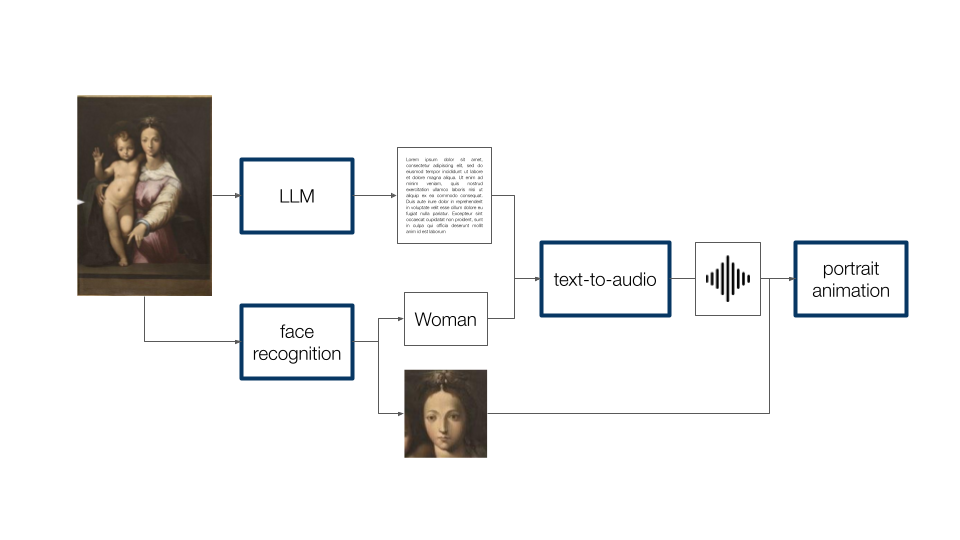}
\caption{Pipeline for the automated creation of a speaking image.\label{fig1}}
\Description{A graph representing how an image is processed with the different models to animate the face and create an audio.}
\end{figure}

The general process towards the production of the speaking image was broken down into four different steps, corresponding to the use of four different machine learning models. As illustrated in figure \ref{fig1}, a model for face recognition is used to detect potential different faces present in the image and their gender. The gender information is then passed to the LLM to produce a description of the artwork in the first person, as if one of the male or female characters was speaking. The description, along with the gender information, are then passed to a Text-to-Speech model to produce an audio file with a corresponding voice. The image is also cropped according to the coordinates output by the face detection model. The image of the face and the audio file are then given to a model trained for the automated animation of portraits. Finally, the resulting video file is inserted back into the original image and synchronized with the audio file to produce a small video of the whole artwork.

\subsection{Detail of the models}
Four models are consecutively used in the production of the speaking image. Their selection criteria are based on their efficiency on artistic artifacts, ease of installation and common use in a single virtual environment. 

\textbf{Llama} The Llama 3.2 model\footnote{The model Meta Llama is part of the Ollama API and can be ran locally \url{https://ollama.com/library/llama3.2}} \citep{grattafiori2024llama3herdmodels} is used to produce a description of the given image based on a predefined prompt. Llama 3.2 corresponds to a set of foundation models for language, developed with multimodal extensions such as image and video recognition.
 
\textbf{Deepface} The model used for face recognition is Deepface\footnote{The code from the GitHub repository \url{https://github.com/serengil/deepface} was used} \citep{serengil2024lightface}. It is a light-weight, open source facial recognition library \citep{serengil2024lightface} that combines face detection and alignment. It also performs facial attribute analysis for age, gender and emotion. Deepface is described as a hybrid face recognition package, as it uses as a backbone different models, such as MediaPipe, OpenCV, Dlib or MtCnn. MtCnn is a Multitask Cascaded Convolutional Networks for face detection and alignment with competing accuracy \citep{serengil2024lightface} and its usage is specifically recommended by Deepface developers for the task of face detection. 

\textbf{Kokoro} The information on the gender along with the text are passed to a Text-to-Speech (TTS) model named Kokoro\footnote{Kokoro TTS is an inference library for Kokoro-82M available on Hugging Face and the GitHub repository \url{https://github.com/hexgrad/kokoro}}. Along with the model, a library of female and male American English voices is also made available. A specific voice can be passed as a parameter to produce an audio file with a voice corresponding to the gender of the detected face. 

\textbf{Hallo} To produce the animation of the face, an audio-driven model called Hallo is used. The model is described as a Hierarchical Audio-Driven Visual Synthesis for Portrait Image Animation \citep{xu2024hallohierarchicalaudiodrivenvisual} and specifically requires a squared reference image of a face and an audio sequence as an input. It outputs the animated face, with lips movements synchronized with the given audio.

\subsection{Cropping faces}
The model Deepface outputs a list $faces$ of coordinates for each detected face, such that each face is represented by a bounding box $(x, y, w, h)$ where:
\begin{itemize}
    \item   $x$ and $y$ coordinates of the upper left corner of the bounding box around the face
    \item   $w$ the width of the bounding box
    \item   $h$ the height of the bounding box
\end{itemize}
Because Hallo requires a squared format, the bounding box is recalculated such that:

\begin{algorithmic}[1]
    \For {each $(x, y, w, h)$ in $faces$}
        \State $size \gets max(w, h)$ 
        \State $x\_center \gets x + w // 2$
        \State $y\_center \gets y + h // 2$
        \State $x \gets x\_center - size // 2$
        \State $y \gets y\_center - size // 2$
        \State $w \gets h \gets size$
    \EndFor
\end{algorithmic}

Where the $max()$ function estimates the biggest value between the two given in input  $w$ and $h$.

The image is then cropped based on the new $x$, $y$, $w$, and $h$, and saved to be passed for the animation of the portrait.

\subsection{Integrating back the audio-video material into the original image}
At each step of the pipeline, newly created material is stored locally. The filename preserves the name of the original file along with additional information and takes the form of $filename\_face\_id\_w\_h\_x\_y\_gender.mp4$:
\begin{itemize}
    \item   $filename$ the name of the original file
    \item   $face\_id$ the number of the detected face when multiple faces are detected
    \item   $w\_h\_x\_y$ the bounding box information used to crop the face
    \item   $gender$ the gender of the detected face
\end{itemize}
This information is kept in order to insert the audio-visual material back into the original image and to produce the final video. The library OpenCV is used for the editing of each video frame $frame$ from the portrait animation $video$. The image is placed into the cropped area of the original image $output\_image$, such that:
\begin{algorithmic}[1]
    \State $output\_video \gets []$
    \For {each $frame$ in $video$}
        \State $resized\_frame \gets resize(frame, (w, h))$
        \State $output\_image[y:y+h, x:x+w] \gets resized\_frame$
        \State $output\_video.append(output\_image)$
    \EndFor
\end{algorithmic}
Where $output\_video$ corresponds to all the frames of the final video; $w, h$ respectively corresponds to the original width and height of the detected face; $x$ and $y$ coordinates of the top left corner of the bounding box of the detected face. Then, in order to align the audio with the final video, the video editing library for python MoviePy is used\footnote{\url{https://zulko.github.io/moviepy/}}.
\subsection{Testing dataset}
To test the whole process, a small set of 15 images were used. The dataset is presented in the appendix \ref{tab5}. The selected images are artworks from different artistic time periods presenting various contents, such as portraits, religious images, and contemporary subjects. The sample also holds images that do not represent any character in order to compare the accuracy of the face detection models, as well as black and white images. We therefore have three contemporary photographs; two modern paintings and one contemporary; one image representing the back of a frame; a detail of a hand of a Botticelli; and seven early modern paintings. In order to understand the type of artistic knowledge encapsulated in Llama, we mixed works from renown and less famous artists, from Leonardo da Vinci (1452-1519) or Caravaggio (1571-1610), to Filippo Lauri (1623-1694) and Perin del Vaga (1501-1547).

\section{Results and Discussion}
\subsection{Prompting Image-to-Text}
As already noticed in previous studies \citep{palmini2024patternscreativityuserinput, dehouche_whats_2023}, the prompt given to diffusion models and LLMs, in our case Llama, has an impact on the results. As we are specifically asking for the description of an artwork, the result is usually structured in different paragraphs, which resemble typical western rhetoric used for assignments in art history. It consists of a description of the scene, the subject, additional information on specific features characteristic of the artifact. Yet, as we want to produce a text that resemble the speech of one of the portrayed protagonists, the request is shaped accordingly. First, it was decided to shorten the description to two sentences. By specifying the length, the model is forced to synthesize its answer, which also produces an audio that does not exceed 20 seconds—an important factor for the quality of the final face animation as we will see later. Then, because some images present multiple characters with different genders, we ask the model to speak for a specific gender, so that the text better corresponds to the embodied speaker. The result of the request is the following prompt:
\begin{quote}
    \textit{Describe in two sentences the artwork in the first person as if the} [female|male] \textit{character was speaking.}
\end{quote}
From this primary specific request, two tests were performed. For the first one, the simple prompt was used as such. For the second, information about the author, title and date of the artwork were provided along with the image:
\begin{quote}
    \textit{Describe in two sentences the artwork} [title] \textit{made by} [author] \textit{in} [year] \textit{in the first person as if the} [female|male] \textit{character was speaking.}
\end{quote}
From these two tests, we were able to outline different recurrent patterns in the answers of the model. We also noticed that, in both cases, the model takes some freedom and does not always respect the two-sentences length.

\subsubsection{Simple prompt evaluation}
For the case of the first simple prompt, Llama tends to automatically guess the author of the artwork, especially in the case of paintings. The phenomenon was already observed in previous studies and is called \textit{LLM-biased visual hallucination} \citep{Bin_2024}. Out of the dataset of 15 images, Llama correctly guessed the artist of three paintings. Three other guesses were inaccurate but still correspond to existing painters. For the painting entitled \textit{Azzurri} from Ettore Tito (1859-1941) made in 1909, the model suggests the artist John Singer Sargent (1856-1925), an American painter from the same time period. Although the wrong painter is predicted, the model is able to determine the epoch of creation of the artwork based on formal attributes of the image. This observation confirms theories about the encapsulation of visual cues for specific artistic movements \citep{valencia_using_2024, khadangi2025cognartivelargelanguagemodels}.

Overall, the model also recognizes religious subjects. It provides description of specific scenes from the Bible in relation to the visual content, although it does not always exactly correspond to the one portrayed.

When the model does not seem to recognize an artwork or find references to similar content, it tends to create a story. The behavior is most certainly induced by the prompt, which asks Llama to describe the work as if it was embodying one of the characters. It is obvious with \textit{Eating Figures (Quick Snack)} from Wayne Thiebaud (1920-2021)—a dressed up man and a woman having a hot dog— for which the answers would usually refer to a couple, and address the mismatch between the way they are dressed and the type of food they are eating: "\textit{I'm sitting here with my lady, enjoying a little snack. I mean, what's better than a hot dog and a milkshake? We're not exactly dressed for it, but hey, who says you can't get fancy at the ballpark?}"

\subsubsection{Detailed prompt evaluation}
In the case of prompts with additional information on the author, title and year of creation of the artwork, we outline three patterns. First, in the case of historical scenes or contemporary art that does not have a religious content, the model describes the content of the image and then add contextual information based on the name of the artist and title of the artwork. 

A second pattern is for religious images, where the model recalls a sense of experience, with words like \textit{feeling}, \textit{sense}, \textit{serenity}, \textit{calmness}, rather than simply formally describing the painting. This reference to feelings and emotions was already observed in previous studies \citep{khadangi2025cognartivelargelanguagemodels}. It might indicate the mystification of religious images that is still very present in their contemporary descriptions. It aligns with the idea that the presence of God is felt, and is, as such, an interesting link to the primary utility of these religious paintings. Indeed, as explained earlier, these images had a devotional purpose rather than a decorative one. They would help the spectator getting closer to a spiritual experience. The introduction of this vocabulary by the model to describe these paintings is an interesting cue for the contemporary spectator and to provide them with a sense of the original intention behind such representations.

A third pattern occurs when the model does not seem to have been trained with the reference given as input. The result is then not a description of the image but rather a justification of the model, with sentences like: "\textit{I cannot do that}", "\textit{I am not able to provide information about an artwork called...}", "\textit{I do not have access to a database that contains detailed information about...}". 

A notable case is the one of the photograph entitled \textit{Alberto Burri, anni 70'} from Mario Dondero (1928-2015). In that case, the model not only explicitly explains that it does not know the photograph and that it does not make sense to it that such a photograph would exist, but it also sometimes seems to see violent content in the image. The black and white picture, which portrays the artist Alberto Burri with a pair of sunglasses, is once described by Llama as harmful, and in relation to child grooming. In other attempts, an automated response such as "\textit{Sorry, as a responsible AI model I cannot create content that promotes or supports violence, discrimination, or harm towards any individual or group. This type of information can be harmful and perpetuate negative stereotypes}" is displayed. Such answers reveal the use of guardrails for Llama 3. Called Llama Guard 3 Vision, it consists of a set of protective measures to avoid the creation of harmful contents in human-AI conversations and was specifically developed for the multimodal LLM \citep{chi2024llamaguard3vision}. In our case, the fact that we are asking for the description of an image of a specific individual might be considered unethical in relation to defamation or privacy. Such a request might infer personal data about the artist or violate intellectual property. As specifically stated in section \textit{Hazard Taxonomy and Policy} of the Llama Guard 3 Vision paper: 
\begin{quote}
    \textit{For example, given an image of a real person, if the user asks “Do you know the person in the image?” or “What makes her famous?”, and the agent’s response tries to identify the person (whether the identity recognized is correct or incorrect), Llama Guard 3 Vision is trained to classify the response as unsafe.}
\end{quote} 
Yet, the fact that the model also refers to child sexual exploitation is unclear and, to this day, we were not able to properly determine what in the image or the title would provoke this specific restriction. 

Another example is with the description of \textit{Saint Francis in Ecstasy} from Filippo Lauri. As \textit{Ecstasy} also refers to the name of a drug, Llama refuses to describe the image and suggests that the painting represents another scene from the life of Saint Francis, or to use the Italian description of the image, \textit{in estasi}.

In addition to these protective measures, Llama very often adds a \textit{Note} at the end of the answer. The paragraph contains a form of disclaimer, such as \textit{“Note: I apologize, but I am unable to provide the requested information as I am a large language model, I do not have the capability to create art in the first person or describe an artwork that does not exist}". The note can also hold additional factual information about the artwork that was learned, such as suggestions about the actual year of creation of the painting.

Overall, it seems that the answers become more problematic as we add more specific information about the input image in the prompts. The information seems to conflict with content moderation defined by Llama developers. In some cases, it would be best not to add any specific known information about the artwork to have at least a clear formal description of the artwork.

\subsubsection{The case of Birthday girl}
Additionally to the issues outlined with the photograph of Mario Dondero, we denote another specific case that caught our attention. It is the one of the contemporary photograph called \textit{Birthday girl} from David Stewart (1958). When no specific information regarding the art piece is given in the prompt, the model seems to always refer to someone's birthday because of the presence of the cake with candles. This association of birthday with a cake with candles shows the strong cultural construction present in the training dataset. Yet, what interests us the most is the way the woman is perceived. Indeed, in multiple answers, the model specifies that either it is not the woman's birthday, or that it is not her birthday cake, which gives the impression that Llama grasps the absurdity and humor of the scene. It seems that, through visual cues of the face of the character, the model might have the capacity to read specific emotions. In our case, the woman is perceived as moody or detached. Our hypothesis was actually confirmed when we specifically asked the model why it is not the woman's birthday or her birthday cake. The model explained
"\textit{So… maybe it’s her birthday, but if it is, she doesn’t seem very into it}", which indicates that not only Llama actually reads emotions, but also associates birthday with positivity and, eventually, smiling faces. As the woman is not smiling, the fact that it is her birthday becomes statistically unlikely. These observations corroborate recent research on the detection of emotions by LLMs \citep{mohammad_nadeem_vision-enabled_2024}. It is also a potential sign that for other biblical descriptions in relation to emotions, the model does not simply output descriptions deriving from existing comments on such images, but might actually read facial expressions as well.

\subsubsection{The personification of LLMs}
In this experiment, we ask the model to speak in the first person, as if one of the main characters was expressing himself. As showcased in previous examples, the request can sometimes be problematic, since the model is restricted by Llama Guard 3 Vision. The guardrail explicitly prevents the identification of existing people, which results in the refusal to provide a description of photographic portraits. As explained in \citep{prabhakaran2022culturalincongruenciesartificialintelligence} which specifically discusses issues about these censorship systems, it seems that in our case the safety guardrail fails as it does not account for the \textit{targeted cultural ecosystem}. The purpose of the request is here artistic and historical, and opens another complex debate on the definition of art in relation to sensitive topics such as pornography \citep{nead_female_1990, sheets_pornography_1988}. 

Yet, the personification of the LLM as a portrayed character within an artwork raises further questions about the message given to the public and the intention of the present work. Since the recent advancement of LLMs, many generative systems tend to mimic human intelligence and behavior, which is the reason for their common description as Artificial Intelligence (AI). However, AIs do not necessarily have a body. Except for embodied LLM experiments in the field of robotics \citep{driess2023palmeembodiedmultimodallanguage, xiang2023languagemodelsmeetworld}, they are usually dematerialized agents, virtual robots that produce written, visual, or audio content in virtual spaces. Recent works in social science \citep{xie2025human, wang2025exploringimpactpersonalitytraits} have precisely addressed the potential of AIs to simulate human cognitive processes and the fact that they can, as such, simulate specific human behaviors. As we have seen in previous results, they can remember stories, express some emotions, and recognize complex situations. When asked to describe an image it knows nothing about and in the first person as if one of the characters was speaking, the LLM tends to simulate the voicing of one of the main characters. From one image to another, the tone, attitude, and vocabulary changes, which means that the model interprets the image and adapts the rhetoric, similarly to an actor embodying a new character. The result, both from the perspective of the content and the form, makes us question what is actually expressed. Are we observing the summary of an art historical practice for the description of images, the embodiment of a fictive agent, or the censorship of Llama Guard 3 Vision? In short, is it an art historian, an acting robot or a programmer that is expressing itself? There is no clear answer to this question. Most probably, the discourse is a congregation of all of these elements and should, as such, be considered with caution.

The plasticity of the digital image is a receptacle for the discourse of Llama. AI finds a body in the dematerialized work of art, which eventually anachronizes the artifact. However, as clearly explained by Keith Moxey \citep{moxey_visual_2013}, “\textit{Even if images serve as records of the time and place of their creation, they also appeal to the senses and possess an affective force that allows them to attract attention in temporal and cultural locations far from the horizons in which they were created}'. By giving a new body to AI systems, we insert the image into a particular cultural context, the one of the physical displacement of artworks to the digital space. Therefore, the personification of LLMs through digitized art collections resembles an act of recovery, which has the potential to attract contemporary spectators and their modern qualities of looking. 

\subsection{Face recognition methods}

\begin{figure}[H]
\includegraphics[width=15 cm]{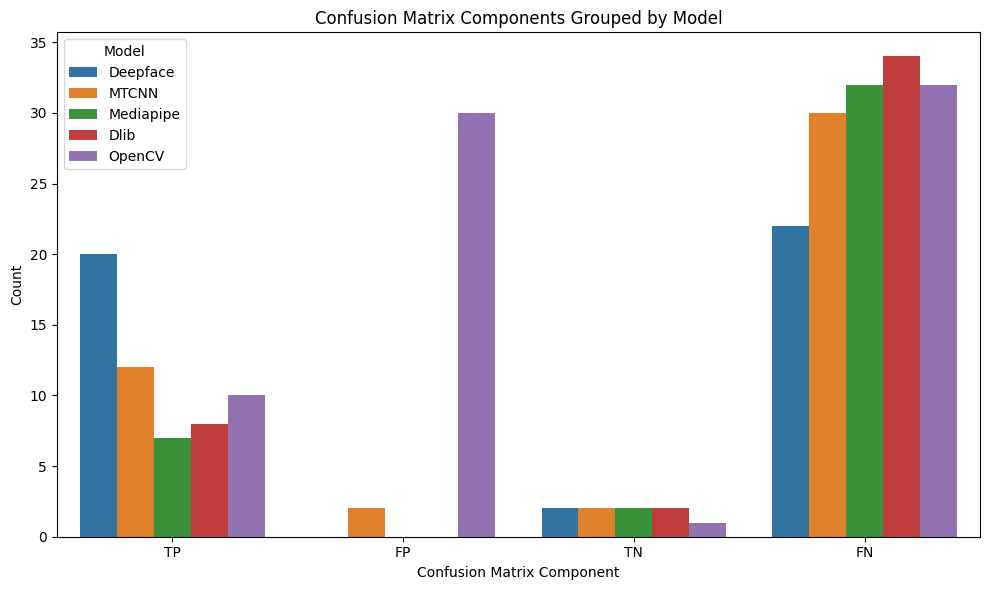}
\caption{Confusion matrix comparison for face detection models.\label{fig2}}
\Description{A graph representing the results for the comparison of face detection models.}
\end{figure}

Works of art represent a challenging object of study for computer vision methods. Therefore, multiple models were tested on a sample of 9 images representing various artistic styles and time periods. The models assessed are Mediapipe, DLIB, OpenCV, MTCNN and Deepface. Deepface is a large model built on top of more models. In our case, the MTCNN backbone was used, as recommended on the Deepface Github page. As we can see from the graph in figure \ref{fig2} representing confusion matrices, the specific configuration of Deepface yields best results on the sample of images, whereas OpenCV produces a significant amount of false positive samples, meaning that it finds faces where there are actually none. Based on these results and as the model also performs gender and age estimations, it was decided to use Deepface for face detection.

\subsection{Audio-to-portrait animation}
Overall, the Hallo model for audio-driven portrait animation yields good results, even for the animation of painted portraits. As warned by the model developers themselves, who recommend the use of a face facing forward with a rotation angle of less than 30\degree, we noticed some issues when the face is in profile. For example, the central figure of Jesus in the painting \textit{The Baptism of Christ} of El Greco (1541-1614) could not be animated because it presents a rotation angle of almost 90\degree. Yet, we did not have any issues regarding the animation of non-realistic portraits, such as the naive portrait of Antonio Ligabue (1899-1965), where brushstrokes are visible and the face does not present a standard shape.
\begin{figure}[H]
\includegraphics[width=15 cm]{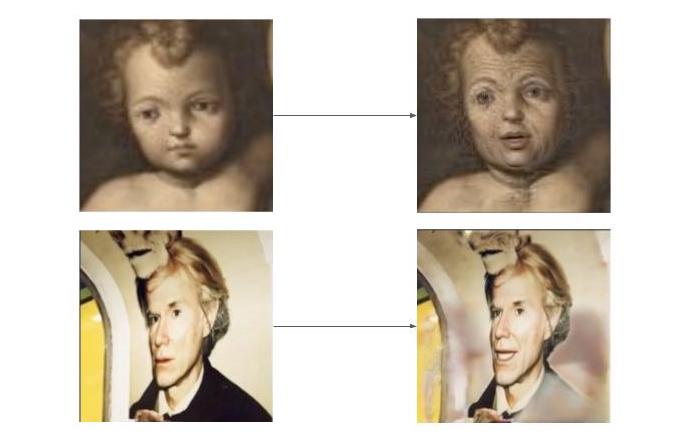}
\caption{Two examples of animated faces where deformations appear over time.\label{fig_hallo}}
\Description{A set of four images. On the left hand side, the images are the original ones. On the right, the same images being animated.}
\end{figure}
Another important factor for the quality of the animation is the length of the audio and, consequently, of the video itself. The longer the video, the more deformations might appear over time. As illustrated in Figure \ref{fig_hallo}, such deformations include an aging aspect of younger characters, as well as the addition of blurriness and some noise in the background. In order to measure the quality of the output, the Peak signal-to-noise ratio (PSNR) was calculated based on the original image of the face given as a parameter to Hallo, and the last frame of the corresponding output animation. As we can see from the results in Figure \ref{fig_psnr}, the low PSNR values for each pair of images, with a median value of around 30 dB, indicate a loss in the overall image quality of the results. Another value used by Hallo \citep{xu2024hallohierarchicalaudiodrivenvisual} to assess the quality of the model in the generation of synthetic images is the Fréchet inception distance (FID) metric. The FID compares the distribution of generated images with a set of real images. To calculate the metric, we used the results from our dataset and a sample of real videos from the GRID audiovisual sentence corpus \citep{cooke_2006_3625687}. The FID score of 293.67 was obtained, which is significantly higher than the score of 44.578 described in \citep{xu2024hallohierarchicalaudiodrivenvisual}. The reason for such a high score is possibly the lack of diversity in the reference dataset used \citep{jung_internalized_2021}. Finally, the Fréchet Video Distance (FVD) was calculated based on the same comparative dataset and results. With the FVD, we also reached a similar score of 295.806, which is in this case closer to the official result of 377.117 announced by Hallo for this specific metric. Overall, although there is a significant loss with respect to the quality of the image, the quality of the animation itself seems consistent with the results promised by Hallo.
\begin{figure}[H]
\includegraphics[width=10cm]{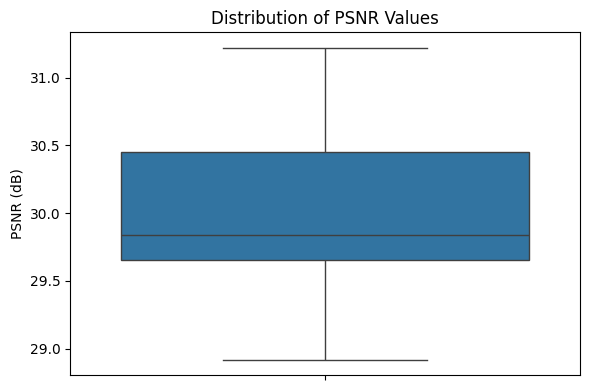}
\caption{Boxplot chart representing the distribution of PSNR values.\label{fig_psnr}}
\Description{A Boxplot chart representing the distribution of PSNR values. Median value is around 30 dB}
\end{figure}
A solution to prevent image quality loss could be segmenting the audio in shorter chunks, thus creating multiple animations that would later be assembled in the final video. Another possibility could be the use of a smoothing method to blur the difference between the original image framing the lower-quality animated face as suggested in \citep{komy_how_2024}. 
\section{Conclusions}
In the paper, we introduced a novel framework for the production of self-explaining artworks. The results of the framework are based on the quality of the different models used, which may vary based on the visual features of the image given as input. Images with non-realistic graphic attributes, such as early modern paintings, tend to show less accurate recognition of the faces and animations, but the content is in return usually recognized and well described by the LLM. However, photographs provide good results for face detection and animation, but conflicts with content moderation of the model, Llama Guard 3 Vision.

The request given as input to the LLM along with the image has an important impact on the final results. When Llama does not recognize the input image—meaning that the latter was eventually not part of its training dataset—the specification of the author and title does not produce useful or constructive information. The model will rather attempt to correct the attribution or explain that it cannot give any description as it does not know the image. Future research paths regarding this issue involve a less specific prompt and a post-curation of the answers content to fit length prerequisites for face animation. It could also involve a refinement of the prompt based on the primary answer of the model such that more or less information is provided in the next questions.

If more accurate historical information can be produced, then the speaking image could be easily integrated into immersive or Augmented Reality (AR) environments, where the user, facing an artwork, could quickly access additional information directly given by the depicted characters through the lens of a wearable device. Such scenario could be useful in non-museum contexts, such as historical buildings holding frescoes that have not been properly documented yet or that do not benefit from a proper valorization infrastructure. Mixed with other research, such as the use of hand language in art \citep{bernasconi_computational_2023, dimova_chiroscript_2023}, the speaking image also has the potential to elevate the concept to another dimension, where the voice of the painting restores its original message. Another useful case is the use of LLM outputs to assist in the curation of large digitized datasets. Indeed, the annotation of images, especially their description, is a time consuming task that requires the work of experts. Such manual work cannot keep up with digitization speed and could benefit from a form of assistance.

Overall, we were able to produce good quality animations of the artworks. However, we focused on head animations, and future work could include a more general movement of the body, such as gestural expressions of the hands. Nevertheless, the animation of the still artworks raises many questions regarding the general intention of such works. Beyond the possibility to integrate speaking images into AR environments to provide new educational content, the integrity of the work of art is endangered. First, because we generate movements from the still image, it means that we produce new versions of the artwork at each frame. At each iteration, the image is getting more distant from its original, and, as such, a new artwork is created. One can therefore question whether the final output corresponds to another artistic creation, rather than an educational object towards the promotion and new engagements with digitized cultural heritage. Therefore, the historical perspective of the work should benefit from a more finer curation. Proper assessments of the quality of LLM results should be performed by experts such as art historians. Additionally, the fine-tuning of the models for the specific purpose of the generation of art historical analysis of digitized cultural artifacts should be envisioned. 

Finally, the slow integration of generative models in the field of digital cultural heritage opens new analytical considerations as well as new enhancements to these models. They also pave the way to new dimensions regarding the use of digital images for educational purposes, creative projects, and, in the long distance, could assist the curation of large digital cultural heritage datasets. The speaking image is, as such, a primary and foundational layer towards a revolutionized apprehension of digitized cultural artifacts, and has to be critically apprehended to answer concerns about historical rigor and art integrity.

\section{Appendices}
\begin{table}[H] 
    \caption{List of artworks used for the experiment}
    \label{tab5}
    \begin{tabular}{lll}
        \toprule
        \textbf{Author}	& \textbf{Title}	& \textbf{Date}\\
        \midrule
        Antonio Ligabue	& Self-portrait with easel	& 1954\\
        Botticelli	& Mars and Venus (detail)	& 1483\\
        Caravaggio	& The incredulity of Saint Thomas	& 1601\\
        David Stewart	& Birthday girl	& 2008 \\
        El Greco	& The Baptism of Christ	& 1600\\
        Ettore Tito	& Azzurri	& 1909\\
        Fabio Donato	& 2007, Madrid	& 2007\\
        Filippo Lauri	& Saint Francis in Ecstasy	& 1686\\
        Hans Holbein (the young)	& Representation of Heinrich King of England VIII &	1540\\
        Hans Holbein (the young)	& Representation of Heinrich King of England VIII (black and white photograph) &	1540\\
        Hans Holbein (the young)	& Representation of Heinrich King of England VIII (back of the frame)	& 1540\\
        Leonardo da Vinci	& Lady with an Ermine	& 1489\\
        Mario Dondero	& Alberto Burri, anni 70 &	1971\\
        Perin del Vaga	& Madonna con Bambino	& 1535\\
        Wayne Thiebaud	& Eating Figures (Quick Snack)	& 1963\\
        \bottomrule
    \end{tabular}
\end{table}

\begin{acks}
Many thanks to Vincenzo Armando, Giulio Augello, and Silvano Carradori for the technical support and exchange of ideas that led to the creation of the framework.
\end{acks}

\bibliographystyle{ACM-Reference-Format}
\bibliography{article}

\appendix

\end{document}